\documentclass{article}
\usepackage{arxiv}
\usepackage[utf8]{inputenc} 
\usepackage[T1]{fontenc}    
\usepackage{url}            
\usepackage{booktabs}       
\usepackage{nicefrac}         
\usepackage{graphicx}
\usepackage[numbers]{natbib}
\usepackage{doi}
\usepackage{makecell}
\usepackage{graphicx}
\usepackage{tikz}
\usepackage{soul}
\usepackage{comment}
\mathchardef\mhyphen="2D
\usepackage{amssymb, amsmath}
\usepackage{color}
\usepackage{multirow}
\usepackage{graphicx,wrapfig,lipsum}
\usepackage{subfigure}
\usepackage{booktabs} 
\usepackage{dsfont}
\usepackage{wrapfig}
\usepackage{mwe}
\usetikzlibrary{backgrounds}
\usepackage{caption}
\usetikzlibrary{intersections}
\usetikzlibrary{shapes.geometric}
\usetikzlibrary{spy}
\usepackage{color, colortbl}
\usepackage{xcolor}
\usepackage[accsupp]{axessibility} 
\definecolor{Gray}{gray}{0.9}
\usepackage[accsupp]{axessibility}  
\usepackage[noend]{algpseudocode}
\usepackage{algorithm}
\usepackage{algorithmicx}
\usepackage{etoolbox}
\usepackage{amsmath}
\usepackage{amsfonts}
\usepackage{txfonts}
\usepackage{xfrac}
\usepackage[noend]{algpseudocode}
\usepackage{chemformula}
\usepackage{nicematrix, mathtools, booktabs, multirow, makecell}

\title{GTFLAT: Game Theory Based Add-On For Empowering Federated Learning Aggregation Techniques}

\date{} 				

\author{{Hamidreza Mahini} \\
	Mälardalen University\\
	\texttt{hamidreza.mahini@gmail.com} \\
	\And
	\hspace{1mm}Hamid Mousavi\\
	Mälardalen University\\
	\texttt{seyedhamidreza.mousavi@mdu.se} \\
	\AND
	\hspace{1mm}Masoud Daneshtalab \\
         Mälardalen University \\
	\texttt{masoud.daneshtalab@mdu.se}
}

\date{}

\renewcommand{\shorttitle}{GTFLAT: Game Theory Based Add-On For Empowering Federated Learning Aggregation Techniques}
\newtheorem{definition}{Definition}
\newtheorem{exmp}{Example}

\newcommand\norm[2]{\left\lVert#1\right\rVert_{#2}}
\newcommand{\RNum}[1]{\textbf{(\uppercase\expandafter{\romannumeral #1\relax})}}

\newcommand{\quotes}[1]{``#1''}
\DeclareMathAlphabet{\mathdcal}{U}{dutchcal}{m}{n}
\SetMathAlphabet{\mathdcal}{bold}{U}{dutchcal}{b}{n}
\DeclareMathAlphabet{\mathdbcal}{U}{dutchcal}{b}{n}
\newcommand{\mypm}{\mathbin{\smash{%
			\raisebox{0.35ex}{%
				$\underset{\raisebox{0.5ex}{$\smash-$}}{\smash+}$%
			}%
		}%
	}%
}
\newcommand\acc[2]{#1$\mypm$ #2}
\newcommand\bacc[2]{\textbf{#1}$\mypm$#2}
\begin{document}
\maketitle
\begin{abstract}
	\texttt{GTFLAT}, as a game theory-based add-on, addresses an important research question: How can a federated learning algorithm achieve better performance and training efficiency by setting more effective adaptive weights for averaging in the model aggregation phase? The main objectives for the ideal method of answering the question are: (1) empowering federated learning algorithms to reach better performance in fewer communication rounds, notably in the face of heterogeneous scenarios, and last but not least, (2) being easy to use alongside the state-of-the-art federated learning algorithms as a new module. To this end, \texttt{GTFLAT} models the averaging task as a strategic game among active users. Then it proposes a systematic solution based on the population game and evolutionary dynamics to find the equilibrium. In contrast with existing approaches that impose the weights on the participants, \texttt{GTFLAT} concludes a self-enforcement agreement among clients in a way that none of them is motivated to deviate from it individually. The results reveal that, on average, using \texttt{GTFLAT} increases the top-1 test accuracy by $1.38$\%, while it needs $21.06$\% fewer communication rounds to reach the accuracy. 
\end{abstract}
\section{Introduction}
\label{sec:introduction}
State-of-the-art Machine Learning (ML) algorithms, notably deep learning, are data-devouring. In the traditional ML, a centralized algorithm works on big centralized data that have been gathered, transmitted, and stored before. While recently, data tend to be distributed at the network's edge considering emerging technologies such as Internet-of-Things (IoT) and new network generations like 5G. Despite generating hundreds of Zettabytes of data mainly produced by edge devices, only a tiny part of them is being stored. On top of that, both reports and forecasts illustrate the share of data centers from the stored data is less than 20\% \cite{GlobalI10:online, Fivethin28:online}. Obviously, there is much data, but not where it used to be. So, ML has come across a new generation of data owners with specific characteristics and requirements. As a prime example, usually, there is no willingness to expose or share local data among them due to privacy issues, escaping from data transmission overhead, or legislative reasons. Therefore, modern ML algorithms have no choice but to be fed on many distributed localized small data instead of centralized stored big data.\\
To this end, \textbf{Federated Learning (FL)} has been proposed as an ML paradigm wanting to reach a trained global model on a server by training several models over remote devices like smartphones or siloed data centers such as hospitals while keeping the clients' data localized \cite{li2020federated}. Generally, there are two scenarios in this context from the scalability perspective: cross-device and cross-silo \cite{bonawitz2022federated}. Both scenarios utilize a central server or service to orchestrate the training process based on their participants' local data. However, they are different from each other in several fundamental views, such as communications, data availability, client reliability, and so forth \cite{kairouz2021advances, zhang2022enabling}.\\
In addition to the commonplace challenges in the traditional ML, there are some new stumbling blocks to reaching good performance in FL. These obstacles mostly come from FL's distributed nature. Heterogeneity is one of the most important things that affect FL performance. The clients can be heterogeneous regarding their resources and, more importantly, their data. The authors paid attention to the impact of resource heterogeneity such as CPU, memory, or network and the data size on training time in FL \cite{232971}. Data heterogeneity is an inescapable fact in FL notably in the cross-device scenarios. Aggregation of clients' models, while they own non-iid (independent and identically distributed) data, is likely to reach a non-optimal global model at the server. So, considering this issue is critical for all proposed FL algorithms \cite{zhu2021federated}.\\ 
The communication cost of uploading and downloading the models is another crucial challenge in FL. Generally, we can address this issue by decreasing the size of transmitted models' parameters or reducing the communication rounds. The number of global rounds and local epochs are two important hyperparameters that control the trade-off between the model's accuracy and training efficiency in FL \cite{morell2022optimising}.\\ 
Another issue is the impact of FL on the environment. Many studies show how much ML algorithms are responsible for producing \ch{CO_2}-equivalents (\ch{CO_2e}). As an illustration, training a large transformer model with Neural Architecture Search (NAS) may produce \ch{CO_2e}, almost 56 times more than the average amount a person is responsible for in a year \cite{strubell2019energy}. Another research investigated this issue in FL and showed that it is more significant than we used to \cite{qiu2021first}. The authors formulated the FL energy consumption and quantified \ch{CO_2e} emissions compared to the centralized ML scenario. Needless to say that the number of global rounds is the most significant parameter affecting the issue. Although the earlier studies had investigated the impact of non-iid data on the required global rounds to achieve the threshold accuracy, they conducted the same experiments to provide more intuition. The results showed that conventional FL algorithms may need to run global rounds of more than 5000 to reach 50\% accuracy for non-iid partitioned \texttt{CIFAR10} data; nearly 4.5 times more than the iid setup. It is tantamount to consuming energy five times and emitting almost 380 times \ch{CO_2e} more than centralized ML.

	\begin{figure}
		\centering
		\includegraphics[width=\linewidth]{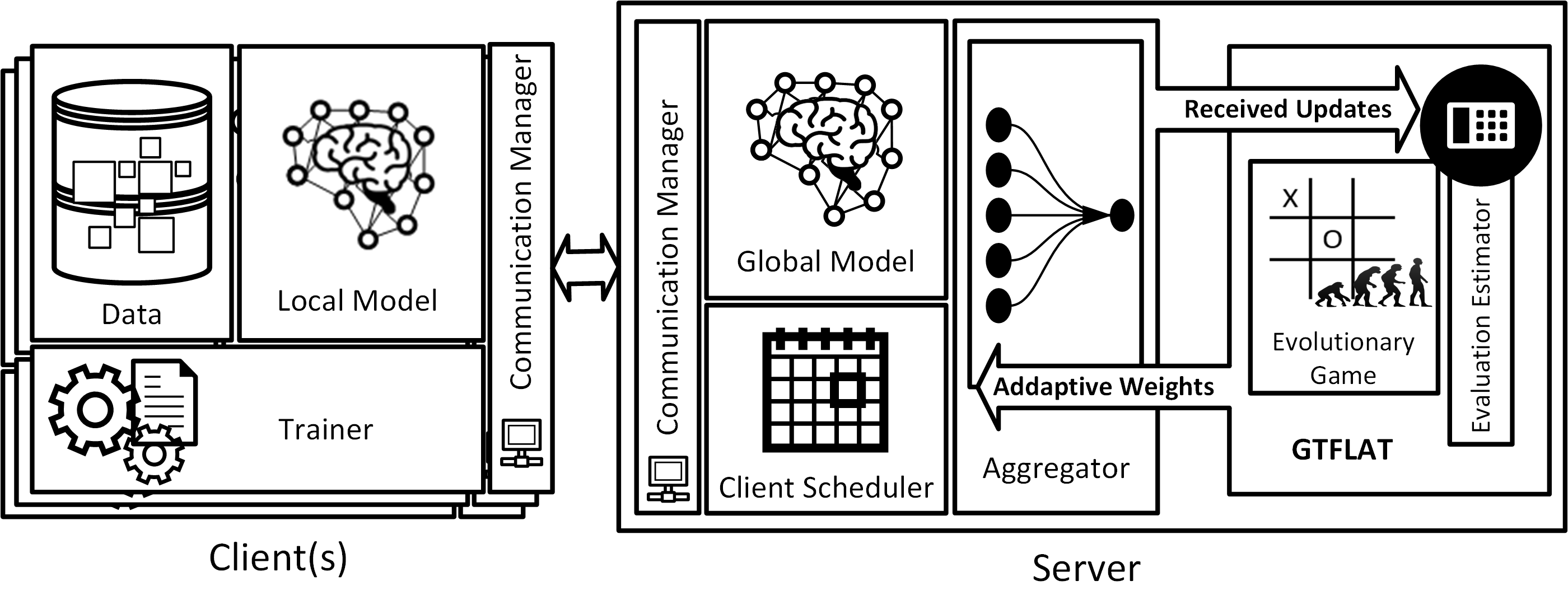}
		\caption{\texttt{GTFLAT} add-on gets the received updates and gives the adaptive weights.}
		\label{fig::gtflat}
	\end{figure}
	
\paragraph{Main Contributions} Carrying the above insight, \texttt{GTFLAT} proposes an add-on for adjusting adaptive weights usable for almost all averaging-based aggregation in FL. Figure \ref{fig::gtflat} depicts the \texttt{GTFLAT}'s overview. In line with the realistic view, the main goal is to reach better accuracy during fewer communication rounds considering non-iid data and fewer active users in each round. Fewer communication rounds and participants, less energy consumption, and fewer \ch{CO2e} emission. Although in the mentioned setting, \texttt{GTFLAT} equips the state-of-the-art FL algorithms so that they stand head and shoulders above what they used to be, it doesn't degrade their performance in other alternate scenarios.\\
Therefore it is possible to summarize our contribution as follows: \RNum{1} We propose a new formulation for the FL aggregation procedure as a strategic game in which the active users play each other to reach an equilibrium as a self-enforcement agreement on the adaptive weights for averaging. \RNum{2} We put forward an estimation based on the distance between each pair of received updates as the game's utility function to avoid imposing extra overhead and new information requirements. \RNum{3} To provide a systematic solution for the proposed game, we devise a mechanism based on population game and evolutionary game principles. The final state of the populations indicates the equilibrium of the game and the averaging weights consequently. \RNum{4} We capsulate the proposed solution as an add-on that gets the received updates as an input and gives the adaptive weights as the output. So, it is possible to use it as a new abstract module in all FL techniques using averaging for the aggregation. \RNum{5} We show \texttt{GTFLAT} improves performance and training efficiency by reaching more accurate models during fewer global FL rounds in heterogeneous scenarios. However, it has no negative impact on the performance in alternate scenarios. We propose a new evaluation parameter called Effective Round Improvement Ratio for more intuition of this claim.
\paragraph{Organization} The rest of the paper is organized as follows.  Section \ref{sec::related_work} provides a brief background about FL and reviews the related work. To establish common terminology, Section \ref{sec::gameback} reviews the game theory and evolutionary dynamics in a nutshell. Section \ref{sec::gtflat_details} discusses the \texttt{GTFLAT} in details. Section \ref{sec::experiments} examines the results of the conducted experiments. Finally, Section \ref{sec::conclusion} concludes the paper and clarifies the research horizon for future work.

\section{FL Background and Related Work}
\label{sec::related_work}
	Formally, FL aims to minimize the empirical risks over local data via running an aggregation algorithm in several global rounds. Equation \ref{eq::flgoal} indicates this fact if there are ${\textstyle m}$ clients and ${\textstyle\mathdcal{L}_k}$ shows the ${\textstyle k}$-th client's objective. Besides, ${\textstyle\omega_k}$ reveals the impact of ${\textstyle k}$-th client on the risk so that ${\textstyle\sum_{k}^{} \omega_k = 1}$ \cite{ji2021emerging}. The majority of FL algorithms aggregate receiving clients' updates via a simple weighted averaging. \texttt{FedAvg} as the vanilla federated averaging algorithm, sets the weights (${\textstyle\omega_k}$) proportional to the number of available data on each participant \cite{mcmahan2017communication}. It is possible to select a portion of clients as the active users in each round of FL. So, Equation \ref{eq::alphak} defines the assigned weight to ${\textstyle k}$-th active user provided that ${\textstyle\mathdcal{A}}$, and ${\textstyle n_k}$ indicate the active users set in a specific round of FL and the number of local training data on ${\textstyle k}$-th user, respectively.\\ 
	
	\begin{equation}
		\label{eq::flgoal}
		\min_{\theta} f(\theta) = \sum_{k=1}^{m}\omega_k\mathdcal{L}_{k}\left(\theta\right)\\	
	\end{equation}
	
	\begin{equation}
		\label{eq::alphak}
		\omega_k = \frac{n_k}{\sum_{i \in \mathdcal{A}}n_i}
	\end{equation}
	
	If ${\textstyle\theta_t^{\left(k\right)}}$, and ${\textstyle\omega_k}$ represent the model's parameters of ${\textstyle k}$-th active user and its weight at round ${\textstyle t}$, respectively, the global model's parameters (${\textstyle\theta_{t+1}}$) are updated based on Equation \ref{eq::flmodelsparam}.  
	
	\begin{equation}
		\label{eq::flmodelsparam}
		\theta_{t+1} = \sum_{k \in \mathdcal{A}} \omega_k \theta_t^{\left(k\right)}
	\end{equation}
	
	Generally, each FL algorithm iterates several global rounds so that each iteration contains four common steps regardless of the employed specific algorithm:
	
	\begin{enumerate}
		\item The server selects a subset of available clients as the active users based on a predefined ratio and a selection strategy. However, uniform random selection is the most prevalent selection strategy among FL algorithms; some recent research mainly focuses on it. This issue usually reputes to client selection \cite{cho2022towards}, participant selection \cite{9714733}, or client scheduling \cite{9530450} in this research context.
		\item The server broadcasts the global model's shared parameters to all active users. Technically, the majority part of algorithms has to share the global model's parameters among clients totally, while there are some efforts such as \texttt{FedGen} to share them partially \cite{zhu2021data}. 
		\item The server waits to receive updates from all active users. Generally, in traditional FL algorithms, the server must wait to receive all updates; However, due to some practical issues such as the heterogeneity based on clients' abilities in communication and computation, relying on synchronous algorithms significantly affects the performance. So, the asynchronous FL algorithms have been proposed in some recent studies \cite{nguyen2022federated}.
		\item The server aggregates the received updates based on the employed FL algorithm and updates the global model's parameters accordingly. It is not an exaggeration if this step is considered as the heart of FL algorithm. As mentioned before, weighted averaging is the most prevalent approach in this phase. In contrast with \texttt{FedAvg}, the weights can be set adaptively in each round \cite{yeganeh2020inverse, chen2019communication}.    
	\end{enumerate}
	
	Algorithm \ref{alg::generalFL} shows these steps in the form of pseudocode.\\
	
	\begin{algorithm*}
		\caption{Federated Learning Server Training}
		\begin{algorithmic}[1]
			
			\Procedure{Server.Train}{Config} \Comment{Config is a struct of all configurations such as the number of FL rounds and so forth}
			\State $M_{\text{global}} \gets$ \Call{InitializeGlobalModel}{Config.Model}
			\For{$i=1$ \textbf{to} Config.FLRounds} 
			\State $U_{\text{active}} \gets$ \Call{SelectActiveUsers}{Config.SelectionStrategy, Config.ActiveUsersRatio} 
			\State \Call{SendModel}{$M_{\text{global}}$, $U_{\text{active}}$} \Comment{broadcast the global model to all active users}
			\ForAll{$U$ \textbf{in}  $U_{\text{active}}$}
			\Call{U.Train}{Config.UserTrainingHyperParams} \Comment{train all active users based on their local data}
			\EndFor
			\While{\textbf{Not} ReceivingAllUpdates}
			\Call{WaitForRecieveingUserUpdate}{ }
			\EndWhile
			\State $M_{\text{global}} \gets$ \Call{Aggregate}{RecievedUpdates} 
			\EndFor
			\State \textbf{return} $M_{\text{global}}$
			\EndProcedure
			
		\end{algorithmic}
		\label{alg::generalFL}
	\end{algorithm*}
	%
	%
	Regarding the aggregation of clients' models, \texttt{FEDAVG} employs an element-wise averaging aggregation scheme over client model weights.
	However, if the client's data is sampled non-iid from their aggregated data, which typically occurs in practice, its performance can suffer due to a drift in the client updates. \cite{khaled2020tighter}.  
	To address this issue, many improvements have been proposed to FEDAVG. 
	From a local training perspective, a few strategies have been put forth to address the heterogeneity issue.
	\texttt{SCAFFOLD} \cite{karimireddy2020scaffold} tries to correct drift in the client's updates by estimating the difference between the server model's and client models' update direction. 
	\texttt{FEDPROX} \cite{li2020federated} makes local training more consistent by including a proxy term in the cost function to prevent model drifts in clients.
	\texttt{FEDBN}\cite{li2021fedbn} uses the local batch normalization layers in the client models and updates them locally without communication and layer aggregation.
	\texttt{FEDENSEMBLE} \cite{shi2021fed} brings model ensembling instead of  aggregating local models to update a single global model in heterogeneous settings.   
	\texttt{FEDGEN} \cite{zhu2021data} utilizes a data-free knowledge distillation approach to learn a generative model and broadcast it to the clients to escort their model training. 
	Another aim is to improve the efficiency of model aggregation by using adaptive weighting and Bayesian approaches. 
	\texttt{IDA} \cite{yeganeh2020inverse} proposed a novel aggregation method that determines the coefficient weights in model aggregation based on the statistical meta-information from model parameters.
	\texttt{FEDMA} \cite{wang2020federated}  created a global shared model by iterative layer-wise matching and averaging hidden components in neural networks with related features. However, it is not suitable for deep models.
	\texttt{FedDF}  uses knowledge distillation in the server and enhances the global model with the ensemble knowledge from local models.  
	\texttt{FEDBE} \cite{chen2020fedbe} uses a bayesian approach to sample better base models and integrates ensemble learning for much more powerful aggregation.
	\texttt{GTFLAT} as a game theory-based add-on helps all FL algorithms that use averaging as the aggregation to reach better accuracy during fewer communication rounds in heterogeneous scenarios. Besides, it guarantees not to affect homogeneous ones negatively. So, the outcome of using \texttt{GTFLAT} leads to Pareto improvement and dominates not using it.

\section{Game theory background}
	\label{sec::gameback}
	Game theory as a mathematical tool studies the models of interactions between \textbf{intelligent} and \textbf{rational} agents (players) \cite{myerson2013game}. Rationality dictates that each player chooses the best strategy based on its intelligence to maximize its profit in a game. Intelligence means that each player has the required knowledge about the game and the adequate resources to interpret the game's situations as the same as a game theorist or designer \cite{narahari2014game, myerson2013game}.\\
	Normal/Strategic form is one of the most common standard game representations, including three elements, a set of players (${\textstyle N}$), a set of available strategies or actions for each player (${\textstyle S_i}$), and the utility function (${\textstyle U_i}$) which is shown in Equation \ref{eq::NormalForm}.
	\begin{equation}
		\label{eq::NormalForm}
		\Gamma = \left\langle N, (S_i)_{i \in N}, (u_i)_{i \in N}\right\rangle
	\end{equation}
	After modeling and representation, we must solve the game. The game solution seeks to intelligently prognosis the game to recommend rational strategies to players. In other words, a solution must define the best response of each player to the selected strategies by the others. Therefore, finding a strategy profile in which all players has no profitable motivation to deviate is a solution for a game. This is one of the most common approaches to solve a game. A strategy profile that all its elements are the best response is a Nash Equilibrium (NE). We call a strategy pure or mixed, depending on it contains only one action or a combination of actions. A mixed strategy is a probability distribution over actions. Therefore, ${\textstyle\sigma_i}$ or, more precisely, ${\textstyle\sigma_i^{(p_1, p_2, p_3,..., p_k)}}$ shows a mixed strategy for player ${\textstyle i}$ who has ${\textstyle k}$ actions. Actions with a non-zero probability of choice in a mixed strategy are called the support set of this strategy and are shown by ${\textstyle\delta(\sigma_i)}$. Moreover, ${\textstyle\Delta(S_i)}$ defines the Set of all probability distributions on ${\textstyle S_i}$. Regarding pure and mixed strategies, a game may have a Pure Strategy Nash Equilibrium (PSNE) or a Mixed Strategy Nash Equilibrium (MSNE). However, MSNE is a superset of PSNE. Definition \ref{def::mixedNE} defines MSNE formally. Note, ${\textstyle S_{-i}}$ indicates the strategies of all player except ${\textstyle i}$.
	\begin{definition}{\textbf{Mixed Strategy Nash Equilibrium (MSNE)}}
		\label{def::mixedNE}
		In a normal form game such as ${\textstyle\Gamma}$, a strategy profile ${\textstyle(\sigma_1^*, \sigma_2^*, ..., \sigma_n^*)}$ is a mixed strategy Nash equilibrium if ${\textstyle\forall i \in N}$,
		\begin{equation}
			\label{eq::mixedNE}
			u_i(\sigma_i^*, \sigma_{-i}^*) \geq u_i(\sigma_i, \sigma_{-i}^*) \; \forall \sigma_i \in \Delta(S_i).
		\end{equation}
	\end{definition}
	\subsection{Population games and evolutionary dynamics}
	\label{subsec::popGame}
	John Nash proved the existence of at least one mixed-strategy Nash equilibrium in every finite strategic form game. A finite game refers to a game with a finite number of players who have finite actions and definitive payments. This is brilliant but not constructive proof. Therefore, finding Nash equilibrium may be like finding the needle in the haystack. Generally, this problem is classified as a PPAD (Polynomial Parity Arguments on Directed graphs) complexity class member that is a subset of NP class \cite{nisan2007algorithmic}. On the other hand, due to inadequate resources, the intelligence principle may not be satisfied in some situations. Due to incomplete computation of the optimal strategy, even fully rational and intelligent players may make a myopic decision based on their rule-of-thumb behavior. This inevitable fact is called bounded rationality in game theory\cite{glazer2016models}. Despite these limitations, if there is an equilibrium in a game and players play the game repeatedly with learning about previous actions, they will modify their strategies and finally reach an equilibrium \cite{myerson2013game}. In other words, players' strategies can converge to equilibrium by a flexible rule-of-thumb and adaptive behavior. This idea implements via two approaches, \textbf{evolutionary game theory} and \textbf{learning in the game}.\\
	Darwin's classic founded the theory of evolution, and John Maynard Smith introduced the evolutionary game theory in the 1980s in his book \cite{smith1982evolution} after his seminal research in the 1970s \cite{smith1972game, smith1973logic}.  The main idea of the evolutionary game lies in a simple natural rule. Every organism selects its strategy inherently based on its inner nature, and a better-fitted strategy leads to more living opportunities and more offspring. Therefore, nature reinforces better strategies, and after a while, organisms select the best responses, and the population converges to an equilibrium. Only one important point of this simple cycle remains: although children inherently follow their parent's strategy, there is always the possibility of a change in strategy due to the mutation. Therefore, population games and using evolutionary dynamics propose another solution for a game that can overcome the mentioned limitations \cite{sandholm2010population}. In such games, players' masses consider as a population with initialized selected strategies. The set of chosen strategies by population is called the state of the game. The initial state can equal a pure or a mixed strategy, and it does not rely on rationality and intelligence. A revision protocol based on a fitness function determines the population's life expectancy, reproduction, and mutation.\\
	There are some specialized terms and definitions in the evolutionary game theory. Using dynamics is a prevalent approach to solving a population game. Regarding dynamics, we can use the replicator equation as defined in Definition \ref{def::RD}.
	\begin{definition}{\textbf{Replicator Dynamics}}
		\label{def::RD}	
		Assume that  there are ${\textstyle k}$ actions such as ${\textstyle s = 1, 2, ..., k}$  in a game. Let us ${\textstyle X_s}$ shows the fraction of the population who are playing ${\textstyle s}$, and  ${\textstyle u(s, \sigma)}$ indicates the expected fitness. Then Equation \ref{eq::RD} defines the replicator equation, which ${\textstyle\sigma(t)}$ is the vector of ${\textstyle X_s(t)}$'s and ${\textstyle\bar{u}}$ equals average fitness at time ${\textstyle t}$.
		\begin{equation}
			\label{eq::RD}
			\begin{split}
				X_s(t+\tau) - X_s(t) &= X_s(t)\frac{\tau\left[u\left(s, \sigma\left(t\right)\right)-\bar{u}\left(\sigma\left(t\right)\right)\right]}{\bar{u}\left(\sigma\left(t\right)\right)}\\
				\bar{u}\left(\sigma\left(t\right)\right)&=\sum_{s=1}^{k}X_s(t)u\left(s, \sigma\left(t\right)\right)
			\end{split}	
		\end{equation}
	\end{definition}
	While the Equation \ref{eq::RD} is valid for any ${\textstyle\tau}$, it gives discrete-time dynamics provided that ${\textstyle\tau = 1}$. Moreover,  the continuous replicator is defined by taking the limit as ${\textstyle\tau \rightarrow 0}$ from Equation \ref{eq::RD} after devding its both sides by ${\textstyle\tau}$. Therefore, Equation \ref{eq::RDC} shows the continuous replicator, that ${\textstyle\dot{X_s} \equiv \sfrac{dX_s(t)}{dt}}$.
	\begin{equation}
		\label{eq::RDC}
		\dot{X_s} = X_s(t)\frac{u\left(s, \sigma\left(t\right)\right)-\bar{u}\left(\sigma\left(t\right)\right)}{\bar{u}\left(\sigma\left(t\right)\right)}
	\end{equation}
	Regarding the aforementioned dynamics definition,  ${\textstyle X^\ast}$ indicates the  \textbf{stationary state} of  Equation \ref{eq::RDC} which means that ${\textstyle\dot{X_{s}^{\ast}} = 0}$ for all ${\textstyle s}$. Furthermore,  dynamics is induced by a neighborhood of ${\textstyle X^\ast}$ with starting from any ${\textstyle X_0}$ that we call it an \textbf{asymptotically stable state}. You can find more details about population games and evolutionary dynamics in \cite{sandholm2010population}.

\section{\texttt{GTFLAT} Details}
\label{sec::gtflat_details}
In a nutshell, \texttt{GTFLAT} sets the adaptive weights of averaging aggregation to improve the FL performance. According to Equation \ref{eq::flmodelsparam} the primary goal is finding the more fit ${\textstyle\omega_i}$ in each round of FL. The global model must satisfy all clients while being accurate enough in the face of unseen data. From the personalization point of view, clients logically tend to the local model trained on their own local data. However, no client can prejudge what happens when it comes across unseen data. Thereby, the server should find an equilibrium profile among clients in which no one motivates to deviate from it. This fact is sufficient to remind us of game theory usage. Hence, we first formulate the mentioned problem as a game and propose a systematic solution based on the principles of evolutionary game theory. The section ends up dissecting \texttt{GTFLAT} approach by an example. 
	
\subsection{Weighted Averaging as a Game}
\label{subsec::assumptions}
In each FL round, \texttt{GTFLAT} server sets up a game among the active users (${\textstyle\mathdcal{A}}$) to calculate the adaptive weights via the obtained equilibrium of the game. Note the equilibrium is interpreted as a self-enforcement agreement among all players, and there is no plus to deviating from it individually. So, according to Equation \ref{eq::NormalForm}, it must define three elements: the set of players (${\textstyle N}$), the set of available actions/strategies for each player (${\textstyle S_i}$), and the utility function ($u_i$). The set of active users as a subset of all existing clients (${\textstyle\mathdcal{C}=\left\{c_0, c_1, \dots c_m\right\}}$) defines the players in the game. Each player has the right to choose the models trained locally in that round apart from its model. As mentioned, every client will likely tend to its model as a dominant strategy. So, the right for self-selection has been eliminated to escape from a strategy profile in which all players select their own model. Let ${\textstyle\mathdcal{M} = \left\{m_0, m_1, \dots m_k\right\}}$ show the set of active users' models. Besides, a desired ML performance criterion (${\textstyle\mathdcal{P}}$) such as model accuracy can define the players' payoffs. Accordingly, Equation \ref{eq::gtflatGame1} represents the proposed game in the strategic form. 
\begin{equation}
	\label{eq::gtflatGame1}
	\left\langle N = \big\{i\mid c_i \in \mathdcal{A}\big\}, S_i = \big\{m_k \mid k \in N - \left\{i\right\} \big\}, u_i \propto \mathdcal{P} \right\rangle
\end{equation}
Perhaps like most game theory-based modelings, defining the utility function is the main stumbling block as the most complex part. It is clear that the more accurate model, the greater its utility. Every player must update and evaluate the model based on each possible strategy profile in a brute force routine. It means imposing a huge communication and computation extra overhead. Adding such a burden is tantamount to the last straw for an intrinsically herculean task such as FL and is not affordable. Hence, the main question is: how does each player can judge the model before updating it? A good Estimation.\\
Let ${\textstyle\varphi_i^j}$ indicate the estimation of player ${\textstyle i}$ from ${\textstyle m_j}$. So, Equation \ref{eq::evalMatrix} defines the matrix of evaluation estimation.
\begin{equation}
\label{eq::evalMatrix}
		\footnotesize
		\varPhi = 
		\begin{pNiceMatrix}[first-row, first-col]
			&m_0			&m_1			&\dots	&m_{k-1}\\
			\addlinespace
			0_{\hspace{.25em}}			&\varphi_0^0	&\varphi_0^1	&\dots	&\varphi_0^{k-1}\\
			\addlinespace
			1_{\hspace{.25em}}			&\varphi_1^0	&\varphi_1^1	&\dots	&\varphi_1^{k-1}\\
			\vdots_{\hspace{.25em}}		&\vdots			&\vdots			&\ddots	&\vdots\\
			\text{\footnotesize $k-1$}	&\varphi_{k-1}^0&\varphi_{k-1}^2&\dots	&\varphi_{k-1}^{k-1}\\
		\end{pNiceMatrix}
\end{equation}
	
Inasmuch as \texttt{GTFLAT} avoids imposing any computation and communication overheads to the clients, the server must run all steps of the game devoid of any extra interaction with them. Thereby, the list of tensors that are received as the updates are all accessible information for the server. A mathematical distance of received updates between each pair of active users can be considered a reasonable estimation to calculate the game's payoff —the less distance, the more excellent utility. Equation \ref{eq::phi} depicts this fact provided that ${\textstyle\theta_i = \left[L_i^0, L_i^1, \dots L_i^{n-1}\right]}$, \texttt{PD} is the abbreviation of \texttt{PairwiseDistance} function and ${\textstyle\norm{x}{F}}$ is the \texttt{Frobenius} norm \cite{gentle2007matrix} of ${\textstyle x}$. It is clear that, by this estimation, ${\textstyle\varPhi}$ is a symmetric matrix that its main diagonal is filled by zero. However, it is possible to consider the other estimations apart from what is defined in this paper to generate the matrix of evaluation estimation.
	\begin{equation}
		\label{eq::phi}
		\varphi_i^j =\begin{cases}
			-\norm{\texttt{PD}\left(\theta_i, \theta_j\right)}{F}, & \text{if }i \ne j\\
			0, & \text{otherwise} 			
		\end{cases} 
	\end{equation}
To complete the modeling, the server must calculate the weights vector (${\textstyle\omega}$) based on players' strategies. Equation \ref{eq::omega} shows how the server can calculate the weights based on the expected value for ${\textstyle|\mathdcal{T}_i|}$. Note ${\textstyle \mathdcal{T}_i}$ indicates the set of players who have chosen ${\textstyle i}$ as their strategy in the game.   
	\begin{equation}
		\label{eq::omega}
		\omega_i = \mathbb{E}\left(|\mathdcal{T}_i|\right)\text{, where }\mathdcal{T}_i=\left\{j \in N\middle| s_j = i\right\}
	\end{equation}
Then, the payoff of each player depends on both ${\textstyle \omega}$ vector and ${\textstyle\varPhi}$ matrix based on Equation \ref{eq::payoff}. Since the order of ${\textstyle\omega}$ is ${\textstyle 1\times k}$ and the ${\textstyle\left(\varPhi_{r, c}\right)_{\substack{0\leq r < k\\ c = i}}}$ is a slice of ${\textstyle\varPhi}$ with ${\textstyle k}$ rows and one column, So, the multiplication of them (${\textstyle u_i}$) is a real number.
	\begin{equation}
		\label{eq::payoff}
		u_i(s_i, s_{-i}) = \omega \times \left(\varPhi_{r, c}\right)_{\substack{0\leq r < k\\ c = i}}
	\end{equation}
So far, we have delineated how we can model FL averaging as a strategic form game. Now, we must present a systematic solution for it. As discussed in Section \ref{sec::gameback}, finding the game's equilibrium is the most prevalent solution in game theory. Generally, there are two approaches to finding it: analytical and simulation. To analyze the game, we must find the best response of each player to the possible other players' strategies. In this regard, ${\textstyle BR_i(s_{-i})}$ shows the best response of ${\textstyle i}$-th player to chosen strategies of the others (all players except ${\textstyle i}$). The server sets up a population game corresponding to the proposed strategic game to provide a systematic solution. It considers there are some populations based on the defined players. Although the available actions for each person are the same as the proposed strategic game, the members of the initial generation choose their strategies inherently and randomly. Then, by considering the obtained outcome and applying a revision protocol, they overhaul their strategies to be more fit than they used to. The server repeats this process for several generations to end up in an asymptotically stable state. Eventually, the server calculates the adaptive weights based on the game's final state. Example \ref{exmp::classicGame} delineates this process vividly.
	
	\begin{exmp}
		\label{exmp::classicGame}
		Assume that there are three clients, and all of them have been active in an FL round. So,  ${\textstyle N=\left\{c_0, c_1, c_2\right\}}$, ${\textstyle S_0=\left\{1, 2\right\}}$, ${\textstyle S_1=\left\{0, 2\right\}}$, and ${\textstyle S_2=\left\{0, 1\right\}}$ consequently. Let ${\textstyle\varPhi}$ be as the matrix below:
		\begin{equation*}
			\footnotesize\varPhi = 
			\begin{pNiceMatrix}[first-row, first-col]
				&m_0	&m_1	&m_2\\
				0_{\hspace{.25em}}	&0		&-0.53	&-0.55\\
				1_{\hspace{.25em}}	&-0.53	&0		&-0.37\\
				2_{\hspace{.25em}}	&-0.55	&-0.37	&0\\
			\end{pNiceMatrix}
		\end{equation*}
		If ${\textstyle s_0=1}$, ${\textstyle s_1=0}$, and ${\textstyle s_2=1}$ are the chosen strategies by the players, according to Equation \ref{eq::omega}, ${\textstyle\omega}$ can be calculated as follows:
		\begin{equation*}
			\small
			\mathdcal{T}_0=\left\{1\right\}, \mathdcal{T}_1=\left\{0, 2\right\}, \mathdcal{T}_2=\left\{\right\} \implies \omega=\begin{pmatrix}\frac{1}{3} &\frac{2}{3} & 0\end{pmatrix}	
		\end{equation*}
		Now we can obtain the players' utility based on Equation \ref{eq::payoff} via multiplying ${\textstyle \omega}$ by ${\textstyle \varPhi}$.
The utility of the players for each strategy profile can be calculated in the same way as it ends up in Table \ref{tab::ex1payoff}. It is clear that because we want to display the payoff matrix for three players, we select one player as a pivot and then demonstrate a separate table for each action of the pivot player (${\textstyle s_0=1}$ and ${\textstyle s_0=2}$). 
		\begin{table}
			\setlength\extrarowheight{2pt}
			\centering
			\caption{The payoff matrix of the game in Example \ref{exmp::classicGame}}
			\small
			\begin{tabular}{|r||c|c|}
				\hline	
				\multicolumn{1}{|c||}{\diaghead(-2,1){aaaaaaa}{1}{2}}&$0 \left(m_0\right)$    & $1 \left(m_1\right)$ \\\hline\hline
				$0 \left(m_0\right)$	& (-0.18, -0.35,-0.49) & (-0.35, -0.18, -0.42)   \\\hline
				$2 \left(m_2\right)$	& (-0.36, -0.30, -0.31) & (-0.54, -0.12, -0.25)   \\\hline
				\multicolumn{3}{c}{$s_0 = 1 \left(m_1\right)$}\\\addlinespace\hline
				\multicolumn{1}{|c||}{\diaghead(-2,1){aaaaaaa}{1}{2}}	&$0 \left(m_0\right)$    & $1 \left(m_1\right)$ \\\hline\hline
				$0 \left(m_0\right)$	& (-0.18, -0.48, -0.37) & (-0.36, -0.30, -0.31)  \\\hline
				$2 \left(m_2\right)$ 	& (-0.37,-0.42, -0.18) & (-0.54, -0.25, -0.12)  \\\hline
				\multicolumn{3}{c}{$s_0 = 2 \left(m_2\right)$}\\	
			\end{tabular}
			\label{tab::ex1payoff}
		\end{table}
As mentioned, we can apply evolutionary game theory to solve the game via simulation. The game's initial state is set in a way that half of each population chooses the first available action, and the other half selects the second one. They revise their strategies over 50 generations, and the final state of the game ends up as follows:
		\begin{equation*}
			\small X = 
			\begin{pNiceMatrix}[first-row, first-col]
				&m_0	&m_1	&m_2\\
				0_{\hspace{.25em}}	& - 	&0.50	&0.50\\
				1_{\hspace{.25em}}	&0.13	& -		&0.87\\
				2_{\hspace{.25em}}	&0.12	&0.88	& -\\
			\end{pNiceMatrix}			
		\end{equation*}
The obtained proportions are tantamount to the expected values for ${\textstyle|\mathdcal{T}_i|}$. So, we can calculate the expected value of ${\textstyle\omega}$ based on Equation \ref{eq::omega}:
		\begin{equation*}
			\small
			\begin{split}		
				&\mathbb{E}\left(|\mathdcal{T}_0|\right)=\frac{0.13+0.12}{3}, \mathbb{E}\left(|\mathdcal{T}_1|\right)=\frac{0.50+0.88}{3}, \mathbb{E}\left(|\mathdcal{T}_2|\right)=\frac{0.50+0.87}{3}\\
				&\implies\omega = \begin{pmatrix}0.08 &0.46 &0.46\end{pmatrix}
			\end{split}
		\end{equation*}
	\end{exmp}


	\section{Experiments}
	\label{sec::experiments}

\begin{figure*}
		\centering $\alpha = 0.05$ (non-iid)\\ 
		\centering \includegraphics[width=\linewidth]{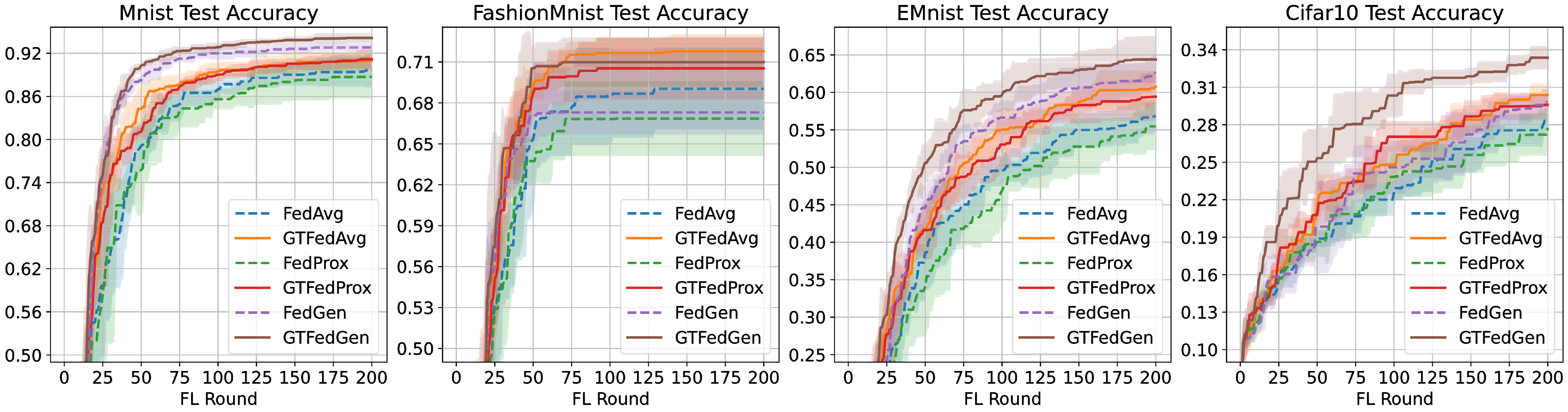}
		\centering $\small\alpha = 1.00$ (almost iid)\\
		\centering \includegraphics[width=\linewidth]{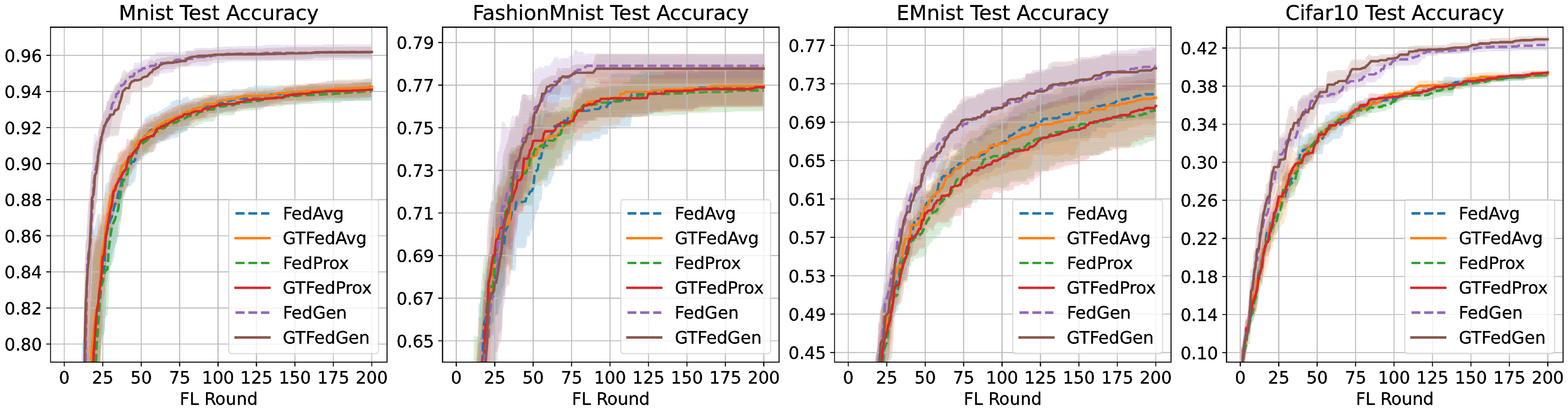}
		\centering
		\caption{The cumulative maximum test accuracy in both non-iid (${\textstyle \alpha=0.05}$) and iid (${\textstyle \alpha=1.00}$) scenarios.}
		\label{fig::results}
	\end{figure*}

 \paragraph{Settings} We apply our experiments on \texttt{CIFAR10} \cite{krizhevsky2009learning}, \texttt{EMNIST} \cite{cohen2017emnist}, \texttt{FashionMNIST} \cite{xiao2017fashionmnist}, and \texttt{MNIST} \cite{lecun-mnisthandwrittendigit-2010} datasets. We imitate some prior studies such as \cite{zhu2021data} in which Dirichlet distribution (${\textstyle \texttt{Dir}(\alpha)}$) has been applied to model non-iid data. The smaller ${\textstyle \alpha}$, the more heterogeneity. In the main setup, we assume there are 50 clients so that 10\% of them participate in each FL round as active users. Random uniform distribution is the employed client selection strategy. The server iterates FL training for 200 rounds. Besides, every active user trains the received model for 20 epochs in each round. To simulate the proposed game, each player starts the game with equal portions of the population for each available strategy. They update their strategies based on Replicator dynamics for 50 generations to reach the final state. We examine the effect of using \texttt{GTFLAT} on \texttt{FedAvg}, \texttt{FedProx}, and \texttt{FedGen} algorithms. 
\paragraph{Results}
We repeat each experiment five times with different random seeds. Table \ref{tab::results} reports the obtained top-1 test accuracy in both non-iid and iid scenarios. We use the prefix \quotes{\texttt{GT}} to illustrate that an FL algorithm has been equipped with \texttt{GTFLAT} add-on. Besides, Figure \ref{fig::results} depicts the cumulative maximum of the test accuracy in all scenarios. To show the performance of \texttt{GTFLAT} vividly, we need the other evaluation parameters beyond test accuracy as follows. 
\begin{definition}{\textbf{Effective Round}}
		\label{def::efr}
		${\textstyle\mathdcal{R}_\texttt{F}^\ast(\mathdcal{P}, \epsilon)}$ defines the minimum required rounds that a particular FL algorithm (${\textstyle \texttt{F}}$) needs to be run to perform as well as ${\textstyle \mathdcal{P}}$ with at most ${\textstyle \epsilon}$ error.
\end{definition}
Furthermore, to compare two FL algorithms, we define \textbf{Effective Round Improvement Ratio} by Equation \ref{eq::erir} provided that ${\textstyle \mathdcal{P}_{\texttt{F}^\prime}^r}$ is equal to the achieved top-1 test accuracy by ${\textstyle \texttt{F}}$ at round ${\textstyle r}$. Positive values for ${\textstyle \mathdcal{E}_{\texttt{F}^\prime}^{\texttt{F}}}$ mean that ${\textstyle \texttt{F}}$ needs fewer rounds to perform as well as ${\textstyle \texttt{F}^\prime}$. In contrast, its negative values show that ${\textstyle \texttt{F}}$ requires more iterations to be as accurate as ${\textstyle \texttt{F}^\prime}$. Table \ref{tab::erir} illustrates the effective round improvement of the mentioned FL algorithms when they employ \texttt{GTFLAT} in both iid and non-iid scenarios.
\begin{equation}
		\label{eq::erir}
		\mathdcal{E}_{\texttt{F}^\prime}^{\texttt{F}}\left(r\right) = \left(1 -  \frac{\mathdcal{R}_\texttt{F}^\ast(\mathdcal{P}_{\texttt{F}^\prime}^r, \epsilon)}{\mathdcal{R}_{\texttt{F}^\prime}^\ast(\mathdcal{P}_{\texttt{F}^\prime}^r, \epsilon)}\right) \times 100
\end{equation}
To make a long story short, results reveal that in the best cases, using \texttt{GTFLAT} increases top-1 accuracy by about 4\% and decreases required communication rounds by more than 52\%; this is while it has just about 0.61\% and 6.72\% negative impacts on the mentioned criteria in the worst cases. On average, it helps the mentioned FL algorithms reach 1.38\% more top-1 test accuracy in 21.06\% fewer communication rounds than they used to.      
	\begin{table*}
		\centering
		\caption{Top-1 test accuracy in both non-iid (${\textstyle \alpha=0.05}$) and iid (${\textstyle \alpha=1.00}$) scenarios at FL round 200 ($\mathdcal{P}_{\texttt{F}}^{200}$).}
		\small
		\begin{tabular}{@{}lcccccccc@{}}
			\toprule
			\multirow{2}{*}{\texttt{Algoritm}}			& 
			\multicolumn{2}{c}{\texttt{MNIST}}          & 
			\multicolumn{2}{c}{\texttt{FashionMNIST}}   & 
			\multicolumn{2}{c}{\texttt{EMNIST}}         & 
			\multicolumn{2}{c}{\texttt{CIFAR10}}  
			\\ \cmidrule(l){2-9} 
			
			& $\alpha=0.05$		& $\alpha=1.00$		& $\alpha=0.05$		&$\alpha=1.00$		& $\alpha=0.05$		&$\alpha=1.00$		& $\alpha=0.05$		& $\alpha=1.00$ \\ 
			\midrule
			\texttt{FedAvg}     & \acc{89.65}{2.05} & \acc{94.11}{0.48}	& \acc{69.04}{2.64}	& \acc{76.88}{0.72} & \acc{56.88}{1.76}	& \bacc{72.15}{2.35}& \acc{28.32}{0.80}	& \acc{39.16}{0.25} \\
			\texttt{GTFedAvg}   & \bacc{91.01}{1.27}& \bacc{94.27}{0.43}& \bacc{71.78}{1.08}& \bacc{76.96}{0.84}& \bacc{60.76}{1.85}& \acc{71.54}{2.50}	& \bacc{30.38}{0.79}& \bacc{39.24}{0.22}\\
			\midrule
			\texttt{FedProx}    & \acc{88.71}{2.63}	& \acc{94.01}{0.43}	& \acc{66.85}{2.41}	& \acc{76.74}{0.84}	& \acc{55.46}{2.73}	& \acc{70.22}{2.62}	& \acc{27.74}{0.82}	& \acc{39.28}{0.14} \\
			\texttt{GTFedProx}  & \bacc{91.17}{1.11}& \bacc{94.11}{0.35}& \bacc{70.53}{1.99}& \bacc{76.90}{0.81}& \bacc{59.43}{2.77}& \bacc{70.70}{2.52}& \bacc{29.59}{1.06}& \bacc{39.44}{0.14}\\
			\midrule
			\texttt{FedGen}     & \acc{92.82}{0.42}	& \acc{96.18}{0.35}	& \acc{67.30}{1.43} & \bacc{77.90}{0.49}& \acc{62.62}{2.07}	& \bacc{74.85}{1.74}& \acc{29.82}{0.59}			& \acc{42.35}{0.44} \\
			\texttt{GTFedGen}   & \bacc{94.14}{0.55}& \bacc{96.19}{0.24}& \bacc{70.99}{1.57}& \acc{77.78}{0.57}	& \bacc{64.40}{2.70}& \acc{74.62}{1.79}& \bacc{33.37}{0.79}			& \bacc{42.92}{0.23}\\ 
			\bottomrule
		\end{tabular}
		\label{tab::results}
	\end{table*}
	\begin{table*}
		\centering
		\caption{Effective Round Improvement Ratio for conducted tests at round 200 and $\epsilon = 5\mathrm{e}{-3}$ (0.5\%) in the non-iid (${\textstyle \alpha=0.05}$) and iid (${\textstyle \alpha=1.00}$) scenarios. To find $\mathdcal{P}_{\texttt{F}}^{200}$ please refer to Table \ref{tab::results}.}
		\small
		\begin{tabular}{@{}lcccrccrccrc@{}}
			\toprule
			\multicolumn{1}{c}{Dataset}	& 
			$\alpha$ &
			\multicolumn{1}{c}{$\mathdcal{R}_\mathtt{FedAvg}^\ast$}	&
			\multicolumn{1}{c}{$\mathdcal{R}_{\texttt{GTFedAvg}\texttt{F}}^\ast$} & 
			\multicolumn{1}{c}{$\mathdcal{E}_{\texttt{FedAvg}}^{\texttt{GTFedAvg}}$}	& 
			
			\multicolumn{1}{c}{$\mathdcal{R}_\mathtt{FedProx}^\ast$}	&
			\multicolumn{1}{c}{$\mathdcal{R}_{\texttt{GTFedProx}\texttt{F}}^\ast$} & 
			\multicolumn{1}{c}{$\mathdcal{E}_{\texttt{FedProx}}^{\texttt{GTFedProx}}$}	& 
			
			\multicolumn{1}{c}{$\mathdcal{R}_\mathtt{FedGen}^\ast$}	&
			\multicolumn{1}{c}{$\mathdcal{R}_{\texttt{GTFedGen}}^\ast$} & 
			\multicolumn{1}{c}{$\mathdcal{E}_{\texttt{FedGen}}^{\texttt{GTFedGen}}$}	& 
			\\ \midrule
			\texttt{MNIST}
			& \multirow{4}{*}{\rotatebox[origin=c]{90}{0.05}} 
			& 159 & 93 & $+41.51\%$		& 150 & 85	& $+43.33\%$	& 131 & 76  & $+41.98\%$\\
			\texttt{FashionMNIST}&	& 101 & 48 & $+52.48\%$		& 70  & 44  & $+37.14\%$	& 47  & 40  & $+14.89\%$\\
			\texttt{EMNIST}&	    & 191 & 127& $+33.51\%$ 	& 193 & 113 & $+41.45\%$	& 197 & 128 & $+35.03\%$\\
			\texttt{CIFAR10}&      	& 196 & 136& $+30.61\%$		& 199 & 128 & $+35.68\%$	& 189 & 94  & $+50.26\%$\\\cmidrule{3-12}
			\textbf{Average}&		&\multicolumn{3}{r}{\textbf{$+$39.53\%}}
			&\multicolumn{3}{r}{\textbf{$+$39.40\%}}
			&\multicolumn{3}{r}{\textbf{$+$35.54\%}}\\ \midrule\midrule	
			
			\texttt{MNIST}
			& \multirow{4}{*}{\rotatebox[origin=c]{90}{1.00}} 
			& 117 & 111 & $+5.13\%$		& 128 & 119 & $+7.03\%$		& 71  & 75  & $-5.63\%$ \\
			\texttt{FashionMNIST}& 	& 113 & 93	& $+17.70\%$	& 98  & 93  & $+5.10\%$		& 73  & 73  & $0.00\%$\\
			\texttt{EMNIST}&       	& 178 & 190 & $-6.72\%$		& 191 & 183 & $+4.19\%$		& 181 & 189 & $-4.42\%$\\
			\texttt{CIFAR10}&      	& 158 & 147 & $+6.96\%$ 	& 171 & 160 & $+6.34\%$		& 152 & 134 & $+11.84\%$\\\cmidrule{3-12}
			\textbf{Average}&	   	&\multicolumn{3}{r}{\textbf{$+$5.77\%}}
			&\multicolumn{3}{r}{\textbf{$+$5.69\%}}
			&\multicolumn{3}{r}{\textbf{$+$0.45\%}}\\
			\bottomrule					
		\end{tabular}
		\label{tab::erir}
	\end{table*}

\section{Conclusion}
\label{sec::conclusion}
In this paper, we proposed a game theory-based add-on called \texttt{GTFLAT} to find adaptive weights for averaging in the aggregation phase of FL algorithms. As a vividly plus, all FL algorithms that use averaging for model aggregation can be benefited from the proposed new module. \texttt{GTFLAT} modeled averaging task as a strategic game among active users in each round of FL. Then, it provided a systematic solution to find the game's equilibrium based on a population game and evolutionary dynamics. In contrast with other approaches in which the server imposes the weights on the clients, \texttt{GTFLAT} concludes a self-enforcement agreement among clients by the obtained equilibrium. From the theoretical point of view, \texttt{GTFLAT} expects to reach better performance because there is no individual motivation to deviate from an equilibrium strategy profile. Despite this, we conducted comprehensive tests on famous datasets and prevalent FL algorithms to practically examine \texttt{GTFLAT} performance. However, we had to show one piece of a larger picture in this paper. Additionally, we proposed a new evaluation parameter called effective round improvement ratio to analyze the results more intuitively. The obtained results proved that using \texttt{GTFLAT} alongside state-of-the-art FL algorithms is a plus for both model performance and training efficiency, notably in heterogeneous scenarios.\\ Finally, considering the model aggregation as a cooperative game and applying a similar idea with \texttt{GTFLAT} to design a mechanism to improve the client selection strategy are our future directions in line with this research.

\end{document}